\documentclass[11pt,a4paper]{article}
\usepackage[hyperref]{acl2021}
\usepackage{times}
\usepackage{latexsym}
\usepackage{subcaption}

\usepackage{url}

\usepackage{graphicx}
\usepackage{verbatim}
\usepackage{booktabs}       
\usepackage{amsfonts}       
\usepackage{nicefrac}       
\usepackage{amsmath,amsfonts,bm}
\usepackage{hyperref}
\usepackage{tikz}
\usetikzlibrary{shapes,arrows,fit,positioning,arrows.meta,backgrounds,bayesnet}
\usepackage{pgfplots}
\pgfplotsset{compat=newest}

\tikzstyle{myarrow} = [>={Stealth[round]},shorten >=1pt,semithick]
\tikzstyle{block} = [draw, inner sep=0.2cm, node distance=0.6cm, fill=yellow!10, font=\small]
\tikzstyle{m} = [draw, inner sep=0.2cm, node distance=0.6cm, fill=blue!10, font=\small]
\tikzstyle{u} = [draw, inner sep=0.2cm, node distance=0.6cm, fill=green!10, font=\small]
\tikzstyle{xblock} = [draw, inner sep=0.2cm, node distance=0.6cm, fill=red!10, font=\small]
\tikzstyle{gray} = [draw, inner sep=0.2cm, node distance=0.6cm, fill=black!03, font=\small]

\usepackage{float}

\hypersetup{
    colorlinks=false,
    linkcolor=blue,
    filecolor=magenta,
    urlcolor=cyan,
}
\def\m{\mathbf{m}}

\def\u{\mathbf{u}}

\def\acc{\text{acc}}
\def\lang{G}
\def\x{\mathbf{x}}

\aclfinalcopy 

\title{Compositionality Through Language Transmission, using Artificial Neural Networks}
 
\author{{\large \bf Hugh Perkins (hp@asapp.com)} \\
  ASAPP (https://asapp.com) \\
  1 World Trade Center, NY 10007 USA
 }

\begin{document}

\maketitle

\begin{abstract}
We propose an architecture and process for using the Iterated Learning Model ("ILM") for artificial neural networks. We show that ILM does not lead to the same clear compositionality as observed using DCGs, but does lead to a modest improvement in compositionality, as measured by holdout accuracy and topologic similarity. We show that ILM can lead to an anti-correlation between holdout accuracy and topologic rho. We demonstrate that ILM can increase compositionality when using non-symbolic high-dimensional images as input.

\end{abstract}

\section{Introduction}

Human languages are compositional. For example, if we wish to communicate the idea of a `red box', we use one word to represent the color `red`, and one to represent the shape `box`. We can use the same set of colors with other shapes, such as `sphere`. This contrasts with a non-compositional language, where each combination of color and shape would have its own unique word, such as `aefabg`. That we use words at all is a characteristic of compositionality. We could alternatively use a unique sequence of letters or phonemes for each possible thought or utterance.

Compositionality provides advantages over non-compositional language. Compositional language allows us to generalize concepts such as colors across different situations and scenarios. However, it is unclear what is the concrete mechanism that led to human languages being compositional. In laboratory experiments using artificial neural networks, languages emerging between multiple communicating agents show some small signs of compositionality, but do not show the clear compositional behavior that human languages show. \citet{kottur2017natural} shows that agents do not learn compositionality unless they have to. In the context of referential games, \citet{lazaridou2018_refgames} showed that agent utterances had a topographic rho of 0.16-0.26, on a scale of 0 to 1, even whilst showing a task accuracy of in excess of 98\%.

In this work, following the ideas of \citet{kirby2001}, we hypothesize that human languages are compositional because compositional languages are highly compressible, and can be transmitted across generations most easily. We extend the ideas of \citet{kirby2001} to artificial neural networks, and experiment with using non-symbolic inputs to generate each utterance.

We find that transmitting languages across generations using artificial neural networks does not lead to such clearly visible compositionality as was apparent in \citet{kirby2001}. However, we were unable to prove a null hypothesis that ILM using artificial neural networks does not increase compositionality across generations. We find that objective measures of compositionality do increase over several generations. We find that the measures of compositionality reach a relatively modest plateau after several generations.

Our key contributions are:

\begin{itemize}
    \item propose an architecture for using ILM with artificial neural networks, including with non-symbolic input
    \item show that ILM with artificial neural networks does not lead to the same clear compositionality as observed using DCGs
    \item show that ILM does lead to a modest increase in compositionality for neural models
    \item show that two measures of compositionality, i.e. holdout accuracy and topologic similarity, can correlate negatively, in the presence of ILM
    \item demonstrate an effect of ILM on compositionality for non-symbolic high-dimensional inputs
\end{itemize}


\section{Iterated Learning Method}
\citet{kirby2001} hypothesized that compositionality in language emerges  because languages need to be easy to learn, in order to be transmitted between generations. \citet{kirby2001} showed that using simulated teachers and students equipped with a context-free grammar, the transmission of a randomly initialized language across generations caused the emergence of an increasingly compositional grammar. \citet{kirby2008}  showed evidence for the same process in humans, who were each tasked with transmitting a language to another participant, in a chain.

\citet{kirby2001} termed this approach the "Iterated Learning Method" (ILM). Learning proceeds in a sequence of generations. In each generation, a teacher agent transmits a language to a student agent. The student agent then becomes the teacher agent for the next generation, and a new student agent is created. A language $\lang$ is defined as a mapping  $G: \mathcal{M} \mapsto \mathcal{U}$ from a space of meanings $\mathcal{M}$ to a space of utterances $\mathcal{U}$. $\lang$ can be represented as a set of pairs of meanings and utterances $G = \{(m_1,u_1),(m_2,u_2), \dots (m_n,u_n) \}$. Transmission from teacher to student is imperfect, in that only a subset, $G_{\text{train}}$ of the full language space $G$ is presented to the student. Thus the student agent must generalize from the seen meaning/utterance pairs $\{ (m_i, u_i) \mid m_i \in M_{\text{train}, t} \subset \mathcal{M} \}$ to unseen meanings, $\{m_i \mid m_i \in (\mathcal{M} \setminus M_{\text{train, t}}) \}$. We represent the mapping from meaning $m_i$ to utterance $u_i$ by the teacher as $f_T(\cdot)$. Similarly, we represent the student agent as $f_S(\cdot)$ In ILM each generation proceeds as follows:
\begin{itemize}
    \item draw subset of meanings $M_{\text{train},t}$ from the full set of meanings $\mathcal{M}$
    \item invention: use teacher agent to generate utterances $U_{\text{train},t} = \{u_{i,t} = f_T(m_{i}) \mid m_i \in M_{\text{train},t} \}$
    \item incorporation: the student memorizes the teacher's mapping from $M_{\text{train},t}$ to $U_{\text{train},t}$
    \item generalization: the student generalizes from the seen meaning/utterance pairs $G_{\text{train},t}$ to determine utterances for the unseen meanings $M_{\text{train},t}$
\end{itemize}

In \citet{kirby2001}, the agents are deterministic sets of DCG rules, e.g. see Figure \ref{fig:dcg_example}. For each pair of meaning and utterance $(m_i, u_i) \in G_{\text{train},t}$, if $(m_i,u_i)$ is defined by the existing grammar rules, then no learning takes place. Otherwise, a new grammar rule is added, that maps from $m_i$ to $u_i$. Then, in the generalization phase, rules are merged, where possible, to form a smaller set of rules, consistent with the set of meaning/utterance pairs seen during training, $G_{\text{train},t}$. The generalization phase uses a complex set of hand-crafted merging rules.

The initial language at generation $t_0$ is randomly initialized, such that each $u_{t,i}$ is initialized with a random sequence of letters. The meaning space comprised two attributes, each having 5 or 10 possible values, giving a total meaning space of $5^2=25$ or $10^2=100$ possible meanings.

\begin{figure}
    \centering
    \noindent\fbox{%
    \parbox{0.3\textwidth}{%
      \small$$
      S: (a_0, b_3) \rightarrow \text{abc}
      $$
      }%
    }
    \noindent\fbox{%
    \parbox{0.3\textwidth}{%
       \small$$
       S:(x, y) \rightarrow A:y\; B:x
       $$
       $$
       A:b_3 \rightarrow \text{ab}
       $$
       $$
       B:a_0 \rightarrow \text{c}
       $$
       }%
       }
    \caption{Two Example sets of DCG rules. Each set will produce utterance `abc' when presented with meanings $(a_0,b_3)$.}
    \label{fig:dcg_example}
\end{figure}

\begin{table}
\small
\centering
    \begin{tabular}{llllll}
    \toprule
           & $a_0$ & $a_1$ & $a_2$ & $a_3$ & $a_4$ \\
         \midrule
         $b_0$ & qda & bguda & lda & kda & ixcda \\
         $b_1$ & qr & bgur & lr & kr & ixcr \\
         $b_2$ & qa & bgua & la & ka & ixca \\
         $b_3$ & qu & bguu & lu & ku & ixcu \\
         $b_4$ & qp & bgup & lp & kp & ixcp \\
    \bottomrule
    \end{tabular}
    \caption{Example language generated by Kirby's ILM.}
    \label{tab:kirby2001example} 
\end{table}

\citet{kirby2001} examined the compositionality of the language after each generation, by looking for common substrings in the utterances for each attribute. An example language is shown in Table \ref{tab:kirby2001example}. In this language, there are two meaning attributes, $a$ and $b$ taking values $\{a_0,\dots,a_4\}$ and $\{b_0,\dots,b_4\}$. For example, attribute $a$ could be color, and $a_0$ could represent `red'; whilst $b$ could be shape, and $b_3$ could represent `square'. Then the word for `red square', in the example language shown, would be `qu'. We can see that in the example, the attribute $a_0$ was associated with a prefix `q', whilst attribute $b_3$ tended to be associated with a suffix `u'. The example language thus shows compositionality.

\citet{kirby2008} extended ILM to humans. They observed that ILM with humans could lead to degenerate grammars, where multiple meanings mapped to identical utterances. However, they showed that pruning duplicate utterances from the results of the generation phase, prior to presentation to the student, was sufficient to prevent the formation of such degenerate grammars.

\section{ILM using Artificial Neural Networks}


\begin{figure}[htbp]
\centering
    \begin{tikzpicture}[]
    \node (m)      [m]
          {$m$};
    \node (sender) [block,right=of m]
          {sender};
    \node (u)      [u, right=of sender]
          {$u$};

    \draw (m) [->,myarrow] to (sender);
    \draw (sender) [->,myarrow] to (u);
    \node (agent) [fit=(sender),inner ysep=0.4cm, yshift=-0.2cm, inner xsep=0.2cm,draw,label={[label distance=-0.6cm]below:Agent}, rounded corners] {};
    \end{tikzpicture}
\caption{Naive ILM using Artificial Neural Networks
\label{naive_rnn_ilm}}
\end{figure}
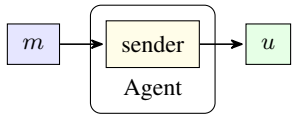

We seek to extend ILM to artificial neural networks, for example using RNNs. Different from the DCG in \citet{kirby2001}, artificial neural networks generalize over their entire support, for each training example. Learning is in general lossy and imperfect.

In the case of using ANNs we need to first consider how to represent a single `meaning'. Considering the example language depicted in Table \ref{tab:kirby2001example} above, we can represent each attribute as a one-hot vector, and represent the set of two attributes as the concatenation of two one-hot vectors.

More generally, we can represent a meaning as a single real-valued vector, $\m$. In this work, we will use `thought vector` and `meaning vector` as synonyms for `meaning`, in the context of ANNs.

We partition the meaning space $\mathcal{M}$ into $M_{\text{train}}$ and $M_{\text{holdout}}$, such that $\mathcal{M} = M_{\text{train}} \cup M_{\text{holdout}} $. We will denote a subset of $M_{\text{train}}$ at generation $t$ by $M_{\text{train}, t}$.

\subsection{Naive ANN ILM}\label{lab:naive_ilm}

A naive attempt to extend ILM to artificial neural networks (ANNs) is to simply replace the DCG in ILM with an RNN, see Figure \ref{naive_rnn_ilm}.



In practice we observed that using this formulation leads to a degenerate grammar, where all meanings map to a single identical utterance. ANNs generalize naturally, but learning is lossy and imperfect. This contrasts with a DCG which does not generalize. In the case of a DCG, generalization is implemented by applying certain hand-crafted rules. With careful crafting of the generalization rules, the DCG will learn a training set perfectly, and degenerate grammars are rare. In the case of using an ANN, the lossy teacher-student training progressively smooths the outputs. In the limit of training over multiple generations, an ANN produces the same output, independent of the input: a degenerate grammar. The first two rows of Table \ref{naive_rnn_ilm} show results for two meaning spaces: 2 attributes each with 33 possible values (depicted as $33^2$), and 5 attributes each with 10 possible values (depicted as $10^5$). The column `uniq' is a measure of the uniqueness of utterances over the meaning space, where 0 means all utterances are identical, and 1 means all utterances are distinct. We can see that the uniqueness values are near zero for both meaning spaces.

We tried the approach of \citet{kirby2008} of removing duplicate utterances prior to presentation to the student. Results for `nodups' are shown in the last two rows of Table \ref{naive_rnn_ilm}. The uniqueness improved slightly, but was still near zero. Thus the approach of \citet{kirby2008} did not prevent the formation of a degenerate grammar, in our experiments, when using ANNs.

\begin{table}
\small
\centering
    \begin{tabular}{lllll}
    \toprule
         Meaning space & Nodups & Uniq & $\rho$ & $\acc_H$ \\
         \midrule
$33^2$ & -  & 0.024 & 0.04 & \textbf{0.05} \\
$10^5$ & -  & 0.024 & 0.08 & 0 \\
$33^2$ & yes  & 0.039 & \textbf{0.1} & 0 \\
$10^5$ & yes  & \textbf{0.05} & \textbf{0.1} & 0 \\
\bottomrule
    \end{tabular}

    \caption{Results using naive ANN ILM architecture. `Nodups': remove duplicates; $\rho$: topographic similarity (see later); `Uniq': uniqueness. Termination criteria for teacher-student training is 98\% accuracy.}
    \label{tab:results_naive}
\end{table}

\subsection{Auto-encoder to enforce uniqueness}\label{lab:autoencoder_ilm}

To prevent the formation of degenerate grammars, we propose to enforce uniqueness of utterances by mapping the generated utterances back into meaning space, and using reconstruction loss on the reconstructed meanings.

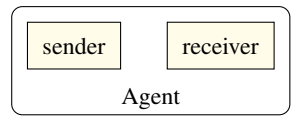
\begin{figure}[ht]
\centering
    \begin{tikzpicture}[auto]
    \node [block] (sender) {sender};
    \node [block] (receiver) [right=of sender] {receiver};
    \node (agent) [fit=(sender) (receiver), yshift=-0.2cm, inner ysep=0.4cm,inner xsep=0.2cm,draw,label={[label distance=-0.5cm]below:Agent}, rounded corners] {};
    \end{tikzpicture}
\caption{Agent sender-receiver architecture
\label{sender_receiver}}
\end{figure}

\begin{figure*}[ht!]
\centering
\begin{subfigure}[t]{0.45\textwidth}
    \centering
    \begin{tikzpicture}[auto]
    \draw[draw=none, use as bounding box] (-0.5, -1.2) rectangle (4, 0.7);
    \node (m)      [m]
          {$m$};
    \node (sender) [block, right=of m]
          {sender};
    \node (u)      [u, node distance=0.7cm, right=of sender]
          {$u$};

    \node (teacher) [fit=(sender), yshift=-0.2cm, inner ysep=0.5cm,inner xsep=0.3cm,draw,label={[label distance=-0.5cm]below:Teacher}, rounded corners] {};

    \draw [myarrow,->] (m) -- (sender);
    \draw [myarrow,->] (sender) -- (u);
    \end{tikzpicture}
    \caption{Teacher sender generates utterances}
\end{subfigure}
\hfill
\begin{subfigure}[t]{0.45\textwidth}
    \centering
    \begin{tikzpicture}[auto]
    \draw[draw=none, use as bounding box] (-0.5, -1.2) rectangle (6, 0.7);
    \node (m)      [m]
          {$m$};
    \node (sender) [block, right=of m]
          {sender};
    \node (u_pred)      [u, node distance=0.7cm, right=of sender]
          {$u_{pred}$};
    \node (u)      [u, node distance=0.7cm, right=of u_pred]
          {$u$};

    \node (student) [fit=(sender), yshift=-0.2cm, inner ysep=0.5cm,inner xsep=0.3cm,draw,label={[label distance=-0.5cm]below:Student}, rounded corners] {};

    \draw [myarrow,->] (m) -- (sender);
    \draw [myarrow,->] (sender) -- (u_pred);
    \draw [<->,dashed,myarrow] (u_pred) -- node [label=below:$\mathcal{L_{CE}}$] {} (u);

    \end{tikzpicture}
    \caption{Train student sender supervised}
\end{subfigure}

\begin{subfigure}[t]{0.45\textwidth}
    \centering
    \begin{tikzpicture}[auto]
    \draw[draw=none, use as bounding box] (-0.5, -1.2) rectangle (6, 0.7);
    \node (m)      [m]
          {$m$};
    \node (m_pred)      [m, right=of m]
          {$m_{pred}$};
    \node (receiver) [block, right=of m_pred]
          {receiver};
    \node (u)      [u, node distance=0.7cm, right=of receiver]
          {$u$};

    \node (student) [fit=(receiver), yshift=-0.2cm, inner ysep=0.5cm,inner xsep=0.3cm,draw,label={[label distance=-0.5cm]below:Student}, rounded corners] {};

    \draw [myarrow,<-] (m_pred) -- (receiver);
    \draw [myarrow,<-] (receiver) -- (u);
    \draw [myarrow,dashed,<->] (m) -- node[label=below:$\mathcal{L_{CE}}$] {} (m_pred);
    \end{tikzpicture}
    \caption{Train student receiver supervised}
\end{subfigure}
\hfill
\begin{subfigure}[t]{0.45\textwidth}
    \centering
    \begin{tikzpicture}[auto]
    \draw[draw=none, use as bounding box] (-1.1, -2.3) rectangle (4.5, 0.5);
    \node (m)      [m]
          {$m$};
    \node (sender) [block, below=of m]
          {sender};
    \node (u)      [u, right=of sender]
          {$u$};
    \node (receiver) [block, right=of u]
          {receiver};
    \node (m_pred)  at (receiver |- m)    [m]
          {$m_{pred}$};

    \node (student) [fit=(sender) (receiver), yshift=-0.2cm, inner ysep=0.5cm,inner xsep=0.3cm,draw,label={[label distance=-0.5cm]below:Student}, rounded corners] {};

    \draw [myarrow,->] (m) -- (sender);
    \draw [myarrow,->] (sender) -- (u);
    \draw [myarrow,->] (u) -- (receiver);
    \draw [myarrow,->] (receiver) -- (m_pred);
    \draw [myarrow,dashed,<->] (m) -- node[label=below:$\mathcal{L}$] {} (m_pred);

    \end{tikzpicture}
    \caption{Train student sender-receiver end-to-end}
\end{subfigure}
\caption{Neural ILM Training Procedure
\label{ilm_procedure}}
\end{figure*}
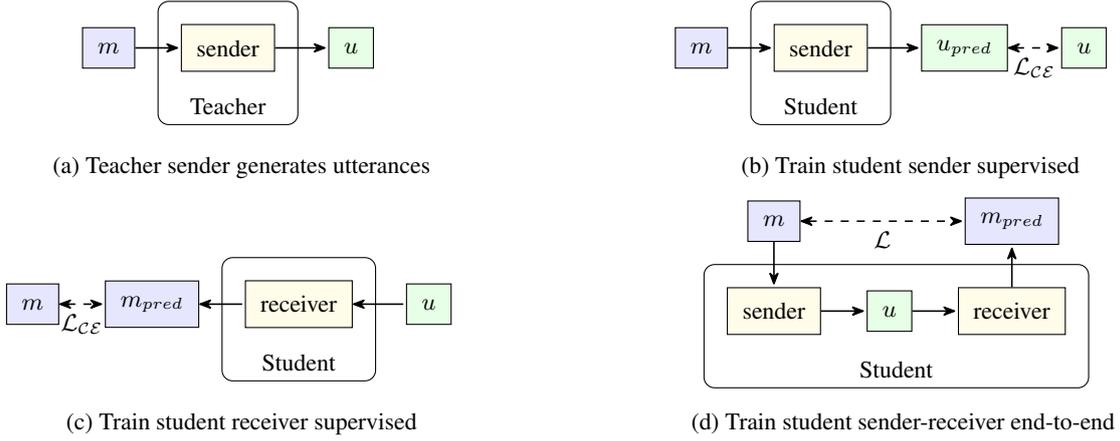

Using meaning space reconstruction loss requires a way to map from generated utterances back to meaning space. One way to achieve this could be to back-propagate from a generated utterance back onto a randomly initialized meaning vector. However, this requires multiple back-propagation iterations in general, and we found this approach to be slow. We choose to introduce a second ANN, which will learn to map from discrete utterances back to meaning vectors. Our architecture is thus an auto-encoder. We call the decoder the `sender', which maps from a thought vector into discrete language. The encoder is termed the `receiver'. We equip each agent with both a sender and a receiver network, Figure \ref{sender_receiver}.

\subsection{Neural ILM Training Procedure}

We will denote the teacher sender network as $f_{T,\text{send}}(\cdot)$, the student receiver network as $f_{S,\text{recv}}(\cdot)$, and the student sender network as $f_{S,\text{send}}$. The output of $f_{\cdot, \text{send}}(\cdot)$ will be non-normalized logits, representing a sequence of distributions over discrete tokens. These logits can be converted into discrete tokens by applying an argmax.

For teacher-student training, we use the sender network of the teacher to generate a set of meaning-utterance pairs, which represent a subset of the teacher's language. We present this language to the student, and train both the sender and the receiver network of the student, on this new language.

The ILM training procedures is depicted in Figure \ref{ilm_procedure}. A single generation proceeds as follows. For each step $t$, we do:

\begin{itemize}
    \item \textbf{meaning sampling} we sample a subset of meanings $M_{\text{train},t} = \{ \m_{t,0}\dots \m_{t,N} \} \subset M_{\text{train}}$, where $M_{\text{train}}$ is a subset of the space of all meanings, i.e. $M_{\text{train}} = \mathcal{M} \setminus  M_{\text{holdout}}$
    \item \textbf{teacher generation}: use the teacher sender network to generate the set of utterances $U_t = \{\u_{t,0}, \dots, \u_{t,N}\}$.
    \item \textbf{student supervised training}: train the student sender and receiver networks supervised, using $M_{\text{train},t}$ and $U_t$
    \item \textbf{student end-to-end training}: train the student sender and receiver network end-to-end, as an auto-encoder
\end{itemize}

For the teacher generation, each utterance $\u_{t,n}$ is generated as $f_{T,\text{send}}(\m_{t,n})$.

For the student supervised training, we train the student receiver network $f_{S,\text{recv}}(\cdot)$ to generate $U_t$, given $M_t$, and we train the student sender network $f_{S,\text{send}}(\cdot)$ to recover $M_t$ given $U_t$. Supervised training for each network terminates after $N_{\text{sup}}$ epochs, or once training accuracy reaches $\acc_{\text{sup}}$

The student supervised training serves to transmit the language from the teacher to the student. The student end-to-end training enforces uniqueness of utterances, so that the language does not become degenerate.

In the end-to-end step, we iterate over multiple batches, where for each batch $j$ we do:
\begin{itemize}
\item sample a set of meanings $M_{\text{train},t,j} = \{ \m_{t,j,0}\dots \m_{t,j,N_{\text{batch}}} \} \subset M_{\text{train}}$
\item train, using an end-to-end loss function $\mathcal{L}_{\text{e2e}}$ as an auto-encoder, using meanings $M_{\text{train}, t,j}$ as both the input and the target ground truth.
\end{itemize}

End-to-end training is run for either $N_{e2e}$ batches, or until end-to-end training accuracy reaches threshold $\text{acc}_{e2e}$

\subsection{Non-symbolic input} \label{lab:non_symbolic_input}

In the general case, the meanings $\m$ can be presented as raw non-symbolic stimuli $\x$. Each raw stimulus $\x$ can be encoded by some network into a thought-vector $\m$. We denote such an encoding network as a `perception' network. As an example of a perception network, an image could be encoded using a convolutional neural network.

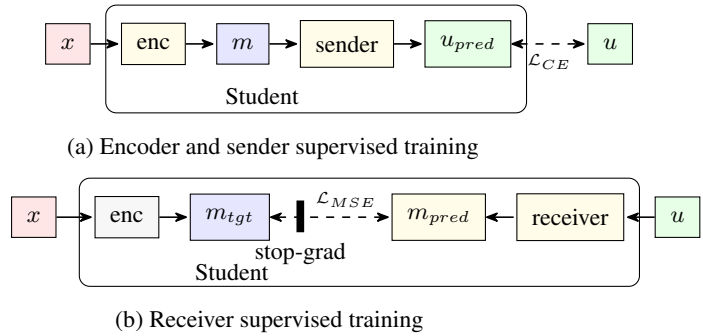
\begin{figure*}[ht]
\centering
\begin{subfigure}[t]{0.4\textwidth}
\begin{tikzpicture}[auto]
    \draw[draw=none, red, use as bounding box] (-0.5, -1.0) rectangle (7.6, 0.7);
    \node [xblock] (x) {$x$};
    \node [block, node distance=0.4cm, right=of x] (encoder) {enc};
    \node [m, node distance=0.4cm, right=of encoder] (m) {$m$};
    \node [block, node distance=0.4cm] (sender) [right=of m] {sender};
    \node [u, node distance=0.4cm, right=of sender] (u_pred) {$u_{pred}$};
    \node [u, node distance=1.0cm, right=of u_pred] (u) {$u$};

    \node (student) [fit=(encoder) (u_pred), yshift=-0.2cm, inner ysep=0.4cm,inner xsep=0.2cm,draw,label={[label distance=-0.5cm]250:Student}, rounded corners] {};

    \draw (x) [->,myarrow] to (encoder);
    \draw (encoder) [->,myarrow] to (m);
    \draw (m) [->,myarrow] to (sender);
    \draw (sender) [->,myarrow] to (u_pred);
    \draw (u_pred) [<->,dashed,myarrow] to node[below,font=\tiny] {$\mathcal{L}_{CE}$} (u);

\end{tikzpicture}
\caption{Encoder and sender supervised training}
\end{subfigure}

\begin{subfigure}[t]{0.45\textwidth}
\begin{tikzpicture}[auto]
    \draw[draw=none, help lines, blue, use as bounding box] (-0.5, -1.0) rectangle (9.0, 0.7);
    \node [xblock] (x) {$x$};
    \node [gray, node distance=0.5cm, right=of x] (encoder) {enc};
    \node [m, node distance=0.4cm, right=of encoder] (m_tgt) {$m_{tgt}$};
    \node [block, node distance=1.6cm] (m_pred) [right=of m_tgt] {$m_{pred}$};
    \node [block, node distance=0.4cm] (receiver) [right=of m_pred] {receiver};
    \node [u, node distance=0.4cm, right=of receiver] (u) {$u$};

    \node (student) [fit=(encoder) (receiver), yshift=-0.2cm, inner ysep=0.4cm,inner xsep=0.2cm,draw,label={[label distance=-1.7cm]195:Student}, rounded corners] {};

    \draw (x) [->,myarrow] to (encoder);
    \draw (encoder) [->,myarrow] to (m_tgt);
    \draw (m_tgt) [<->,dashed,myarrow,font=\tiny] to node [near start,rectangle,solid,draw,minimum height=0.4cm,minimum width=0.08cm, yshift=-0.2cm, inner sep=0, fill=black,label={below:stop-grad}] {}  node[near end,above, xshift=-0.2cm, font=\tiny] {$\mathcal{L}_{MSE}$} (m_pred);
    \draw (m_pred) [<-,myarrow] to (receiver);
    \draw (u) [->,myarrow] -- (receiver);

\end{tikzpicture}
\caption{Receiver supervised training}
\end{subfigure}
\caption{Generalized Neural ILM Supervised Training
\label{fig:generalized_supervised_training}}
\end{figure*}

This then presents a challenge when training a receiver network. One possible architecture would be for the receiver network to generate the original input $\x$. We choose instead to share the perception network between the sender and receiver networks in each agent. During supervised training of the sender, using the language generated by the teacher, we train the perception and sender networks jointly. To train the receiver network, we hold the perception network weights constant, and train the receiver network to predict the output of the perception network, given input utterance $\u$ and target stimulus $\x$. See Figure \ref{fig:generalized_supervised_training}. Note that by setting the perception network as the identity operator, we recover the earlier supervised training steps.

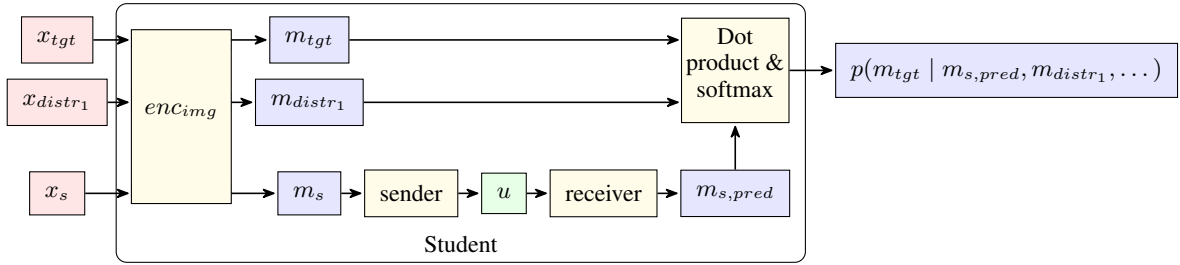
\begin{figure*}[ht]
\centering
\begin{tikzpicture}[auto]

    \node [xblock, left=of encoder] (x_tgt) {$x_{tgt}$};
    \node [xblock, node distance=0.2cm, below=of x_tgt] (x_distr1) {$x_{distr_1}$};
    \node [xblock, below=of x_distr1] (x_s) {$x_s$};
    
    \node [draw, node distance=0.5cm, right=of x_tgt, minimum width=1.3cm] (enc_tgt) {};
    \node [draw, minimum width=1.3cm] at (enc_tgt |- x_distr1) (enc_distr1) {};
    \node [draw, minimum width=1.3cm] at (enc_tgt |- x_s) (enc_s) {};

    \node [m, node distance=0.5cm, right=of enc_tgt] (m_tgt) {$m_{tgt}$};
    \node [m] at (m_tgt |- x_distr1) (m_distr1) {$m_{distr_1}$};
    \node [m] at (m_tgt |- x_s) (m_s) {$m_s$};
    
    \node [block,fit=(enc_tgt) (enc_s), inner sep=0cm] {$enc_{img}$};

    \node [block, node distance=0.3cm, right=of m_s] (sender) {sender};
    \node [u, node distance=0.3cm, right=of sender] (u) {$u$};
    \node [block, node distance=0.3cm, right=of u] (receiver) {receiver};
    \node [m, node distance=0.3cm, right=of receiver] (m_s_pred) {$m_{s,pred}$};

    \node [draw, minimum width=1.4cm] at (m_s_pred |- m_tgt) (dp_tgt) {tgt};
    \node [draw, minimum width=1.4cm] at (m_s_pred |- m_distr1) (dp_distr1) {d1};

    \node [block,font=\small,fit=(dp_tgt) (dp_distr1), inner sep=0cm] (dp) {Dot product \& softmax};
    
    \node [m,font=\small, right=of dp] (p_m_tgt) {$p(m_{tgt} \mid m_{s,pred}, m_{distr_1}, \dots) $};

    \draw (x_tgt) [->,myarrow] -- (enc_tgt);
    \draw (x_distr1) [->,myarrow] -- (enc_distr1);
    \draw (x_s) [->,myarrow] -- (enc_s);

    \draw (enc_tgt) [->,myarrow] -- (m_tgt);
    \draw (enc_distr1) [->,myarrow] -- (m_distr1);
    \draw (enc_s) [->,myarrow] -- (m_s);

    \draw (m_s) [->,myarrow] -- (sender);
    \draw (sender) [->,myarrow] -- (u);
    \draw (u) [->,myarrow] -- (receiver);
    \draw (receiver) [->,myarrow] -- (m_s_pred);

    \draw (m_tgt) [->,myarrow] -- (dp_tgt);
    \draw (m_distr1) [->,myarrow] -- (dp_distr1);
    \draw (m_s_pred) [->,myarrow] -- (dp);

    \draw (dp) [->,myarrow] -- (p_m_tgt);

    \node (student) [fit=(encoder) (enc_tgt) (m_s_pred), yshift=-0.2cm, inner ysep=0.4cm,inner xsep=0.2cm,draw,label={[label distance=-0.5cm]below:Student}, rounded corners] {};

\end{tikzpicture}
\caption{End-to-end referential task for non-symbolic inputs, where $x_s$ is the input stimulus presented to the sender, $x_{tgt}$ is the target input simulus, and $x_{distr_1}$ is a distractor stimulus.
\label{referential_architecture}}
\end{figure*}

For end-to-end training, with non-symbolic input, we use a referential task, e.g. as described in \citet{lazaridou2018_refgames}. The sender network is presented the output of the perception network, $\m$, and generates utterance $\u$. The receiver network chooses a target image from distractors which matches the image presented to the sender. The target image that the receiver network perceives could be the original stimulus presented to the sender, or it could be a stimulus which matches the original image in concept, but is not the same stimulus. For example, two images could contain the same shapes, having the same colors, but in different positions. Figure \ref{referential_architecture} depicts the architecture, with a single distractor. In practice, multiple distractors are typically used.

\subsection{Discrete versus soft utterances}



When we train a sender and receiver network end-to-end, we can put a softmax on the output of the sender network $f_{\cdot, \text{send}}$, to produce a probability distribution over the vocabulary, for each token. We can feed these probability distributions directly into the receiver network $f_{\cdot, \text{recv}}$, and train using cross-entropy loss. We denote this scenario \textsc{softmax}.

Alternatively, we can sample discrete tokens from categorical distributions parameterized by the softmax output. We train the resulting end-to-end network using REINFORCE. We use a moving average baseline, and entropy regularization. This scenario is denoted \textsc{rl}.

\subsection{Evaluation of Compositionality}

We wish to use objective measures of compositionality. This is necessary because the compositional signal is empirically relatively weak. We assume access to the ground truth for the meanings, and use two approaches: topographic similarity, $\rho$, as defined in \citet{brighton_kirby_2006_topographic_mappings} and \citet{lazaridou2018_refgames}; and holdout accuracy $\acc_H$.

$\rho$ is the correlation between distance in meaning space, and distance in utterance space, taken across multiple examples. For the distance metric, we use the $L_0$ distance, for both meanings and utterances. That is, in meaning space, the distance between `red square` and `yellow square` is 1; and the distance between `red square' and `yellow circle' is 2. In utterance space, the difference between `glaxefw' and `glaxuzg' is 3. Considered as an edit distance, we consider substitutions; but neither insertions nor deletions. For the correlation measure, we use the Spearman's Rank Correlation.

$\acc_H$ shows the ability of the agents to generalize to combinations of shapes and colors not seen in the training set. For example, the training set might contain examples of `red square', `yellow square', and `yellow circle', but not `red circle'. If the utterances were perfectly compositional, both as generated by the sender, and as interpreted by the receiver, then we would expect performance on `red circle' to be similar to the performance on `yellow circle'. The performance on the holdout set, relative to the performance on the training set, can thus be interpreted as a measure of compositionality.

Note that when there is just a single attribute, it is not possible to exclude any values from training, otherwise the model would never have been exposed to the value at all. Therefore $\acc_H$ is only a useful measure of compositionality when there are at least 2 attributes.

We observe that one key difference between $\rho$ and $\acc_H$ is that $\rho$ depends only on the compositional behavior of the sender, whereas $\acc_h$ depends also on the compositional behavior of the receiver. As noted in \citet{lowe2019pitfalls}, it is possible for utterances generated by a sender to exhibit a particular behavior or characteristic without the receiver making use of this behavior or characteristic.





\section{Related Work}

Work on emergent communications was revived recently for example by \citet{lazaridou2016multi} and \citet{foerster_multi_agent_rl_2016}. \citet{mordatch2018emergence} and \citet{leibo2017multi} showed emergent communications in a 2d world. Several works investigate the compositionality of the emergent language. \citet{kottur2017natural} showed that agents do not generate compositional languages unless they have to. \citet{lazaridou2018_refgames} used a referential game with high-dimensional non-symbolic input, and showed the resulting languages contained elements of compositionality, measured by topographic similarity. \citet{bouchacourt_baroni_how_agents_see_things} caution that agents may not be communicating what we think they are communicating, by using randomized images, and by investigating the effect of swapping the target image. \citet{andreas2017translating} proposed an approach to learn to translate from an emergent language into a natural language. Obtaining compositional emergent language can be viewed as disentanglement of the agent communications. \citet{locatello2019challenging} prove that unsupervised learning of disentangled representations is fundamentally impossible without inductive biases both on the considered learning approaches and the data sets.


Kirby pioneered ILM in \citet{kirby2001}, extending it to humans in \citet{kirby2008}. \citet{griffiths_kalish} proved that for Bayesian agents, that the iterated learning method converges to a distribution over languages that is determined entirely by the prior, which is somewhat aligned with the result in \citet{locatello2019challenging} for disentangled representations. \citet{bowling_easeofteaching}, \citet{gupta_deepgenerationaltransmission}, and \citet{kirby2020} extend ILM to artificial neural networks, using symbolic inputs.
Symbolic input vectors are by nature themselves compositional, typically, the concatenation of one-hot vectors of attribute values, or of per-attribute embeddings (e.g. \citet{kottur2017natural}). Thus, these works show that given compositional input, agents can generate compositional output. In our work, we extend ILM to high-dimensional, non-symbolic inputs. However, a concurrent work \citet{dagan2020co} also extends ILM to image inputs, and also takes an additional step in examining the effect of genetic evolution of the network architecture, in addition to the cultural evolution of the language that we consider in our own work.

\citet{andreas_measuring_compositionality_2019} provides a very general framework, \textsc{TRE}, for evaluating compositionality, along with a specific implementation that relates closely to the language representations used in the current work. It uses a learned linear projection to rearrange tokens within each utterance; and a relaxation to enable the use of gradient descent to learn the projection. Due to time pressure, we did not use \textsc{TRE} in our own work.

Our work on neural ILM relates to distillation \citep{distillationcaruana} \citep{distillationhinton}, in which a large teacher networks distills knowledge into a smaller student network. More recently, \citet{bornagainnetworks} showed that when the student network has identical size and architecture to the teacher network, distillation can still give an improvement in validation accuracy on a vision and a language model. Our work relates also to self-training \citep{st2019} in which learning proceeds in iterations, similar to ILM generations.

\section{Experiments}
\begin{table*}[ht!]
\small
\centering
    \begin{tabular}{llllll}
    \toprule
$\mathcal{M}$ & $\mathcal{L}$ & E2E Tgt & ILM? & $\acc_H$ & $\rho$ \\
\midrule
$33^2$ & \textsc{softmax} & e=100k &  & 0.97+/-0.02 & 0.23+/-0.01 \\
$33^2$ & \textsc{softmax} & e=100k & yes & \textbf{0.984+/-0.002} & \textbf{0.30+/-0.02} \\
\midrule
$33^2$ & \textsc{rl} & e=500k &  & 0.39+/-0.01 & 0.18+/-0.01 \\
$33^2$ & \textsc{rl} & e=500k & yes & \textbf{0.52+/-0.04} & \textbf{0.238+/-0.008} \\
\midrule
$10^5$ & \textsc{softmax} & e=100k &  & \textbf{0.97+/-0.01} & 0.22+/-0.02 \\
$10^5$ & \textsc{softmax} & e=100k & yes & 0.56+/-0.06 & \textbf{0.28+/-0.01} \\
\midrule
$10^5$ & \textsc{rl} & e=500k &  & \textbf{0.65+/-0.17} & 0.17+/-0.02 \\
$10^5$ & \textsc{rl} & e=500k & yes & 0.449+/-0.004 & \textbf{0.28+/-0.01} \\
\bottomrule
    \end{tabular}
    \caption{
    Results using auto-encoder architecture on synthetic concepts dataset.  "E2E Tgt": termination criteria ("target") for end-to-end training; "$\rho$": topographic similarity. Where ILM is used, it is run for 5 generations.
    }
    \label{tab:results_autoencoder} 
\end{table*}

\subsection{Experiment 1: Symbolic Input}

\subsubsection{Dataset construction}

We conduct experiments first on a synthetic concept dataset, built to resemble that of \citet{kirby2001}.

We experiment conceptually with meanings with $a$ attributes, where each attribute can take one of $k$ values. The set of all possible meanings $\mathcal{M}$ comprises $k^a$ unique meanings. We use the notation $k^a$ to describe such a meaning space. We reserve a holdout set $\mathcal{H}$ of 128 meanings, which will not be presented during training. This leaves $(k^a - 128)$ meanings for training and validation. In addition, we remove from the training set any meanings having 3 or more attributes in common with any meanings in the holdout set.

We choose two meanings spaces: $33^2$ and $10^5$. $33^2$ is constructed to be similar in nature to \citet{kirby2001}, whilst being large enough to train an RNN without immediately over-fitting. With 33 possible values per attribute, the number of possible meanings increases from $10^2=100$ to $33^2 \approx 1,000$. In addition to not over-fitting, this allows us to set aside a reasonable holdout set of 128 examples. We experiment in addition with a meaning space of $10^5$, which has a total of $100,000$ possible meanings. We hypothesized that the much larger number of meanings prevents the network from simply memorizing each meaning, and thus force the network to naturally adopt a more compositional representation.

\subsubsection{Experimental Setup}


The model architecture for the symbolic concept task is that depicted in Figure \ref{ilm_procedure}.

The sender model converts each meaning into a many-hot representation, of dimension $k \cdot a$, then projects the many-hot representation into an embedding space.

\subsubsection{Results}

Table \ref{tab:results_autoencoder} shows the results for the symbolic concept task. We can see that when using an RL link, ILM improves the topographic similarity measure, for both $33^2$ and $10^5$ meaning spaces. This is true for both \textsc{softmax} and \textsc{rl}. Interestingly, in the $10^5$ meaning space, the increase in compositionality as measured by $\rho$ is associated with a decrease in $\acc_H$, for both \textsc{softmax} and \textsc{rl}. This could indicate potentially that ILM is inducing the sender to generate more compositional output, but that the receiver's understanding of the utterance becomes less compositional, in this scenario. It is interesting that $\rho$ and $\acc_H$ can be inversely correlated, in certain scenarios. This aligns somewhat with the findings in \citet{lowe2019pitfalls}.

Interestingly, it is not clear that using a $10^5$ meaning space leads to more compositional utterances than the much smaller $33^2$ meaning space.

\subsection {Experiment 2: Images}

\subsubsection{Dataset}

In Experiment One, we conserved the type of stimuli used in prior work on ILM, e.g. \citet{kirby2001}, using highly structured input. In Experiment Two, we investigate the extent to which ILM shows a benefit using unstructured high-dimensional input. We used OpenGL to create scenes containing colored objects, of various shapes, in different positions.

In the previous task, using symbolic meanings, we required the listener to reconstruct the symbolic meaning. In the case of images, we use a referential task, as discussed in Section \ref{lab:non_symbolic_input}.  The advantage of using a referential task is that we do not require the agents to communicate the exact position and color of each object, just which shapes and colors are present. If the agents agree on an ordering over shapes, then the number of attributes to be communicated is exactly equal to the number of objects in the images. The positions of the objects are randomized to noise the images. We also varied the colors of the ground plane over each image.




\begin{figure*}[ht!]
\centering
\includegraphics[width=0.68000\textwidth, clip, viewport=14px 8px 721px 356px]{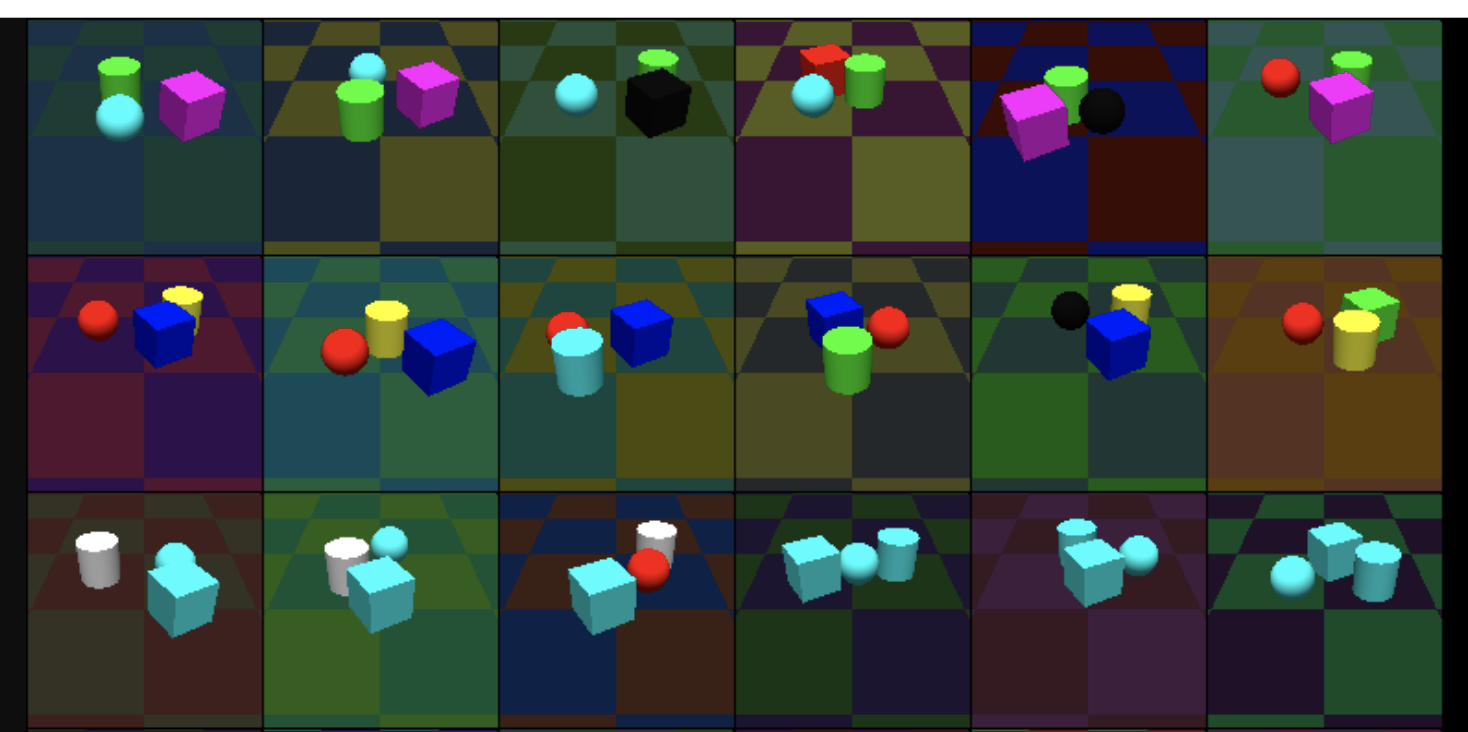}
\caption{Example referential task images, one example per row. The sender image and the correct receiver image are the first two images in each row.
\label{sample_opengl_images}}
\end{figure*}

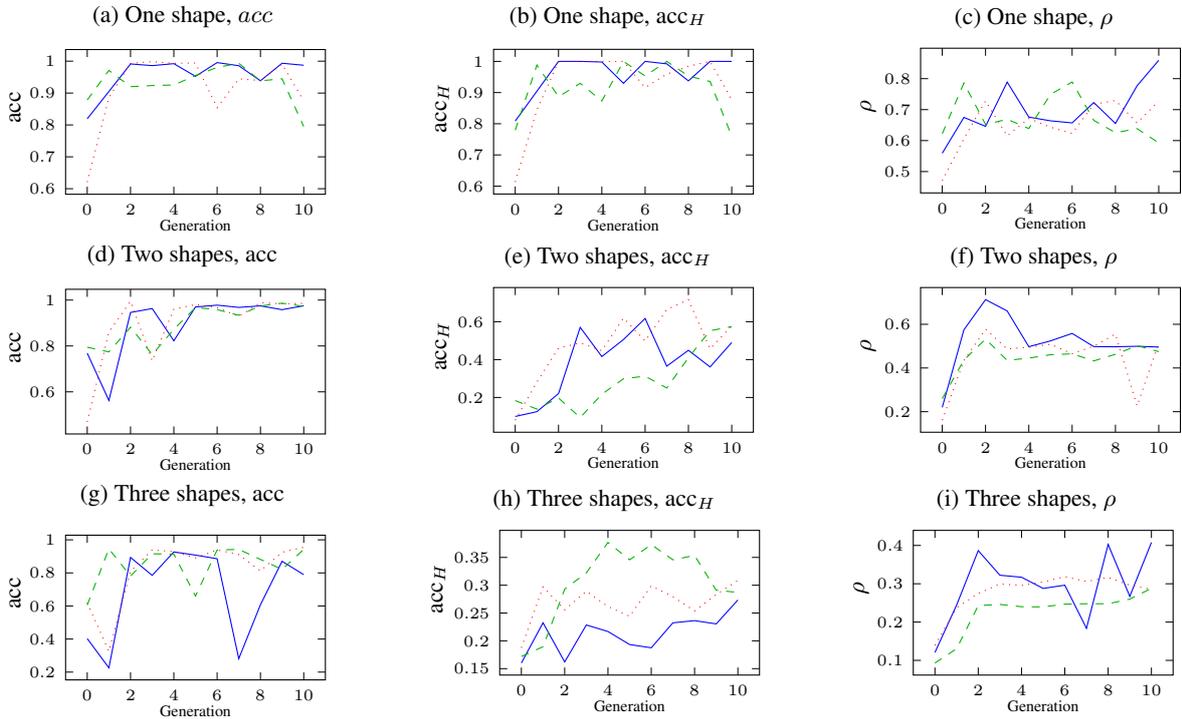
\begin{figure*}[ht!]
\centering
\begin{subfigure}{0.3\textwidth}
    \subcaption{One shape, $acc$}
\begin{tikzpicture}
    \begin{axis}[
        clip=false,
        tiny,
        width=5cm,
        height=3.5cm,
    ]
    \addplot [blue] table [x=generation,y=rt336_e2e_acc, col sep=comma] {data/ilm_runs.csv};
    \addplot [red,dotted] table [x=generation,y=rt332_e2e_acc, col sep=comma] {data/ilm_runs.csv};
    \addplot [black!30!green,dashed] table [x=generation,y=rt331_e2e_acc, col sep=comma] {data/ilm_runs.csv};
    \node [anchor=near xticklabel,font=\tiny,yshift=0.2cm] at (xticklabel cs:0.5) {Generation};
    \node [anchor=near yticklabel,font=\small,xshift=0.15cm,rotate=90] at (yticklabel cs:0.6) {$\acc$};
    \end{axis}
    \end{tikzpicture}
\end{subfigure}
\hfill
\begin{subfigure}{0.3\textwidth}
    \subcaption{One shape, $\acc_H$}
\begin{tikzpicture}
    \begin{axis}[
        clip=false,
        tiny,
        width=5cm,
        height=3.5cm,
    ]
    \addplot [blue] table [x=generation,y=rt336_e2e_holdout_acc, col sep=comma] {data/ilm_runs.csv};
    \addplot [red,dotted] table [x=generation,y=rt332_e2e_holdout_acc, col sep=comma] {data/ilm_runs.csv};
    \addplot [black!30!green,dashed] table [x=generation,y=rt331_e2e_holdout_acc, col sep=comma] {data/ilm_runs.csv};
    \node [anchor=near xticklabel,font=\tiny,yshift=0.2cm] at (xticklabel cs:0.5) {Generation};
    \node [anchor=near yticklabel,font=\small,xshift=0.15cm,rotate=90] at (yticklabel cs:0.6) {$\acc_H$};
    \end{axis}
\end{tikzpicture}
\end{subfigure}
\hfill
\begin{subfigure}{0.3\textwidth}
    \subcaption{One shape, $\rho$}
\begin{tikzpicture}
    \begin{axis}[
        clip=false,
        tiny,
        width=5cm,
        height=3.5cm,
    ]
    \addplot [blue] table [x=generation,y=rt336_e2e_rho, col sep=comma] {data/ilm_runs.csv};
    \addplot [red,dotted] table [x=generation,y=rt332_e2e_rho, col sep=comma] {data/ilm_runs.csv};
    \addplot [black!30!green, dashed] table [x=generation,y=rt331_e2e_rho, col sep=comma] {data/ilm_runs.csv};
    \node [anchor=near xticklabel,font=\tiny,yshift=0.2cm] at (xticklabel cs:0.5) {Generation};
    \node [anchor=near yticklabel,font=\small,xshift=0.15cm,rotate=90] at (yticklabel cs:0.6) {$\rho$};
    \end{axis}
\end{tikzpicture}
\end{subfigure}

\begin{subfigure}{0.3\textwidth}
    \subcaption{Two shapes, $\acc$}
\begin{tikzpicture}
    \begin{axis}[
        clip=false,
        tiny,
        width=5cm,
        height=3.5cm,
    ]
    \addplot [blue] table [x=generation,y=rt333_e2e_acc, col sep=comma] {data/ilm_runs.csv};
    \addplot [red,dotted] table [x=generation,y=rt334_e2e_acc, col sep=comma] {data/ilm_runs.csv};
    \addplot [black!30!green,dashed] table [x=generation,y=rt335_e2e_acc, col sep=comma] {data/ilm_runs.csv};
    \node [anchor=near xticklabel,font=\tiny,yshift=0.2cm] at (xticklabel cs:0.5) {Generation};
    \node [anchor=near yticklabel,font=\small,xshift=0.15cm,rotate=90] at (yticklabel cs:0.6) {$\acc$};
    \end{axis}
\end{tikzpicture}
\end{subfigure}
\hfill
\begin{subfigure}{0.3\textwidth}
    \subcaption{Two shapes, $\acc_H$}
\begin{tikzpicture}
    \begin{axis}[
        clip=false,
        tiny,
        width=5cm,
        height=3.5cm,
    ]
    \addplot [blue] table [x=generation,y=rt333_e2e_holdout_acc, col sep=comma] {data/ilm_runs.csv};
    \addplot [red,dotted] table [x=generation,y=rt334_e2e_holdout_acc, col sep=comma] {data/ilm_runs.csv};
    \addplot [black!30!green,dashed] table [x=generation,y=rt335_e2e_holdout_acc, col sep=comma] {data/ilm_runs.csv};
    \node [anchor=near xticklabel,font=\tiny,yshift=0.2cm] at (xticklabel cs:0.5) {Generation};
    \node [anchor=near yticklabel,font=\small,xshift=0.15cm,rotate=90] at (yticklabel cs:0.6) {$\acc_H$};
    \end{axis}
\end{tikzpicture}
\end{subfigure}
\hfill
\begin{subfigure}{0.3\textwidth}
    \subcaption{Two shapes, $\rho$}
\begin{tikzpicture}
    \begin{axis}[
        clip=false,
        tiny,
        width=5cm,
        height=3.5cm,
    ]
    \addplot [blue] table [x=generation,y=rt333_e2e_rho, col sep=comma] {data/ilm_runs.csv};
    \addplot [red,dotted] table [x=generation,y=rt334_e2e_rho, col sep=comma] {data/ilm_runs.csv};
    \addplot [black!30!green, dashed] table [x=generation,y=rt335_e2e_rho, col sep=comma] {data/ilm_runs.csv};
    \node [anchor=near xticklabel,font=\tiny,yshift=0.2cm] at (xticklabel cs:0.5) {Generation};
    \node [anchor=near yticklabel,font=\small,xshift=0.15cm,rotate=90] at (yticklabel cs:0.6) {$\rho$};
    \end{axis}
\end{tikzpicture}
\end{subfigure}

\begin{subfigure}{0.3\textwidth}
    \subcaption{Three shapes, $\acc$}
\begin{tikzpicture}
    \begin{axis}[
        clip=false,
        tiny,
        width=5cm,
        height=3.5cm,
    ]
    \addplot [blue] table [x=generation,y=rt320_e2e_acc, col sep=comma] {data/ilm_runs.csv};
    \addplot [red,dotted] table [x=generation,y=rt322_e2e_acc, col sep=comma] {data/ilm_runs.csv};
    \addplot [black!30!green,dashed] table [x=generation,y=rt324_e2e_acc, col sep=comma] {data/ilm_runs.csv};
    \node [anchor=near xticklabel,font=\tiny,yshift=0.2cm] at (xticklabel cs:0.5) {Generation};
    \node [anchor=near yticklabel,font=\small,xshift=0.15cm,rotate=90] at (yticklabel cs:0.6) {$\acc$};
    \end{axis}
\end{tikzpicture}
\end{subfigure}
\hfill
\begin{subfigure}{0.3\textwidth}
    \subcaption{Three shapes, $\acc_H$}
\begin{tikzpicture}
    \begin{axis}[
        clip=false,
        tiny,
        width=5cm,
        height=3.5cm,
    ]
    \addplot [blue] table [x=generation,y=rt320_e2e_holdout_acc, col sep=comma] {data/ilm_runs.csv};
    \addplot [red,dotted] table [x=generation,y=rt322_e2e_holdout_acc, col sep=comma] {data/ilm_runs.csv};
    \addplot [black!30!green,dashed] table [x=generation,y=rt324_e2e_holdout_acc, col sep=comma] {data/ilm_runs.csv};
    \node [anchor=near xticklabel,font=\tiny,yshift=0.2cm] at (xticklabel cs:0.5) {Generation};
    \node [anchor=near yticklabel,font=\small,xshift=0.15cm,rotate=90] at (yticklabel cs:0.6) {$\acc_H$};
    \end{axis}
    \end{tikzpicture}
    \end{subfigure}
\hfill
    \begin{subfigure}{0.3\textwidth}
    \subcaption{Three shapes, $\rho$}
    \begin{tikzpicture}
    \begin{axis}[
        clip=false,
        tiny,
        width=5cm,
        height=3.5cm,
    ]
    \addplot [blue] table [x=generation,y=rt320_e2e_rho, col sep=comma] {data/ilm_runs.csv};
    \addplot [red,dotted] table [x=generation,y=rt322_e2e_rho, col sep=comma] {data/ilm_runs.csv};
    \addplot [black!30!green, dashed] table [x=generation,y=rt324_e2e_rho, col sep=comma] {data/ilm_runs.csv};
    \node [anchor=near xticklabel,font=\tiny,yshift=0.2cm] at (xticklabel cs:0.5) {Generation};
    \node [anchor=near yticklabel,font=\small,xshift=0.15cm,rotate=90] at (yticklabel cs:0.6) {$\rho$};
    \end{axis}
\end{tikzpicture}
\end{subfigure}
\caption{Examples of individual ILM runs up to 10 generations.
\label{10gen_all}}
\end{figure*}

Example images are shown in Figure \ref{sample_opengl_images}. Each example comprises 6 images: one sender image, the target receiver image, and 4 distractor images. 
Each object in a scene was a different shape, and we varied the colors and the positions of each object. Each shape was unique within each image. Two images were considered to match if the sets of shapes were identical, and if the objects with the same shapes were identically colored. The positions of the objects were irrelevant for the purposes of judging if the images matched.



We change only a single color in each distractor, so that we force the sender and receiver to communicate all object colors, not just one or two. We create three datasets, for sets of 1, 2 or 3 objects respectively. Each dataset comprises 4096 training examples, and 512 holdout examples.

In the case of two shapes and three shapes, we create the holdout set by setting aside combinations of shapes and colors which are never seen in the training set. That is, the color `red' might have been seen for a cube, but not for a cylinder. In the case of just one shape, this would mean that the color had never been seen at all, so for a single shape, we relax this requirement, and just use unseen geometrical configurations in the holdout set.


The dataset is constructed using OpenGL and python. The code is available at \footnote{https://github.com/asappresearch/neural-ilm}.


\subsubsection{Experimental setup}

The supervised learning of the student sender and receiver from the teacher generated language is illustrated in Figure \ref{fig:generalized_supervised_training}. The referential task architecture is depicted in Figure \ref{referential_architecture}. Owing to time pressure, we experimented only with using \textsc{rl}. We chose \textsc{rl} over \textsc{softmax} because we felt that \textsc{rl} is more representative of the discrete nature of natural languages.

\subsubsection{Results}

\begin{table}[ht]
\small
\centering
    \begin{tabular}{llllll}
    \toprule
Shapes & ILM? & Batches & $\acc_H$ & Holdout $\rho$ \\
\midrule
1 &  &  300k & 0.76+/-0.11 & 0.55+/-0.03 \\
1 & Yes & 300k & \textbf{0.95+/-0.03} & \textbf{0.69+/-0.04} \\
\midrule
2 &  & 600k & 0.21+/-0.03 & 0.46+/-0.2 \\
2 & Yes & 600k & \textbf{0.30+/-0.06} & \textbf{0.64+/-0.05} \\
\midrule
3 &  & 600k & 0.18+/-0.01 & 0.04+/-0.02 \\
3 & Yes & 600k & \textbf{0.23+/-0.02} & \textbf{0.19+/-0.04} \\
\bottomrule
\end{tabular}
    \caption{
    Results for OpenGL datasets. `Shapes' is number of shapes, and `Batches' is total number of batches. For ILM, batches per generation is total batches divided by number of ILM generations. For ILM, three generations are used.}
    \label{tab:results_opengl_01_001} 
\end{table}

Table \ref{tab:results_opengl_01_001} shows the results using the OpenGL datasets. We can see that when training using the \textsc{rl} scenario, ILM shows an improvement across both $33^2$ and $10^5$ meaning spaces.

The increase in topographic similarity is associated with an improvement in holdout accuracy, across all scenarios, similar to the $33^2$ symbolic concepts scenario.


Figure \ref{10gen_all} shows examples of individual runs. The plots within each row are for the same dataset, i.e. one shape, two shapes, or three shapes. The first column shows the end to end accuracy, the second column shows holdout accuracy, $\acc_H$, and the third column shows topologic similarity $\rho$. We note firstly that the variance across runs is high, which makes evaluating trends challenging. Results in the table above were reported using five runs per scenario, and pre-selecting which runs to use prior to running them.

We can see that end to end training accuracy is good for the one and two shapes scenario, but that the model struggles to achieve high training accuracy in the more challenging three shapes dataset. The holdout accuracy similarly falls dramatically, relative to the training accuracy, as the number of shapes in the dataset increases. Our original hypothesis was that the more challenging dataset, i.e. three shapes, would be harder to memorize, and would thus lead to better compositionality. That the holdout accuracy actually gets worse, compared to the training accuracy, with more shapes was surprising to us.

Similarly, the topological similarity actually becomes worse as we add more shapes to the dataset. This seems unlikely to be simply because the receiver struggles to learn anything at all, since the end to end training accuracy stays relatively high across all three datasets. We note that the ILM effect is only apparent over the first few generations, reaching a plateau after around 2-3 generations.






\section{Conclusion}

In this paper, we proposed an architecture to use the iterated learning method (``ILM") for neural networks, including for non-symbolic high-dimensional input. We showed that using ILM with neural networks does not lead to the same clear compositionality as observed for DCGs. However, we showed that ILM does lead to a modest increase in compositionality, as measured by both holdout accuracy and topologic similarity. We showed that holdout accuracy and topologic rho can be anti-correlated with each other, in the presence of ILM. Thus caution might be considered when using only a single one of these measures. We showed that ILM leads to an increase in compositionality for non-symbolic high-dimensional input images.

\section*{Acknowledgements}

Thank you to Angeliki Lazaridou for many interesting discussions and ideas that I've tried to use in this paper.

\bibliography{rnn_ilm_acl}
\bibliographystyle{acl_natbib}

\clearpage
\section*{Appendix: hyper-parameters}

For all experiments, results and error bars are reported using five runs per scenario. We pre-select which runs to use for reporting before running them.

\subsection{Experiment 1}

For experiment 1, we use a batch-size of 100, embedding size of 50. RNNs are chosen to be GRUs. We query the teacher for utterances for 40\% of the training meaning space each generation. We use an utterance length of 6, and a vocabulary size of 4.

\subsection{Experiment 2}

For experiment 2, we use the same architecture as \citet{lazaridou2018_refgames}, with the exception that we add a max pooling layer after the convolutional network layers, with kernel size 8 by 8; and we replace the stride 2 convolutional layers by stride 1 convolutional layers, followed by 2 by 2 max pooling layers. 



We use entropy regularization for both the sender and receiver networks, as per \citet{lazaridou2018_refgames}. At test-time, we take the argmax, instead of sampling.

Other hyper-parameters were as follows:
\begin{itemize}
    \item optimizer: RMSProp
    \item convolutional layers: 8
    \item batch size: 32
    \item no gradient clipping
    \item utterance length: 6
    \item utterance vocabulary size: 100
    \item embedding size: 50
    \item RNN type: GRU
    \item Number RNN layers: 1
    \item dropout: 0.5
    \item supervised training fraction: 0.4
    \item number supervised training steps: 200k
    \item number end to end training steps: 200k
    \item sender entropy regularization: 0.01
    \item receiver entropy regularization: 0.001
\end{itemize}

\end{document}